%% file: iclr2023_conference.tex
\documentclass{article} % For LaTeX2e
\usepackage{iclr2023_conference,times}

\input{math_commands.tex}

\usepackage{hyperref}
\usepackage{url}

\usepackage{times}
\usepackage{latexsym}

\usepackage[T1]{fontenc}
\usepackage[utf8]{inputenc}

\usepackage{microtype}

\usepackage{inconsolata}

\title{Explicit Knowledge Transfer for Weakly-Supervised Code Generation}

\author{Zhangir Azerbayev, Ansong Ni, Hailey Schoelkopf, Dragomir Radev \\
  Yale University \\
  \texttt{zhangir.azerbayev@yale.edu}}
  
\usepackage{amsmath}
\usepackage{amssymb}
\usepackage{graphicx}
\usepackage{mdframed}
\graphicspath{ {./images/} }
\newcommand{\specialcell}[2][c]{%
  \begin{tabular}[#1]{@{}c@{}}#2\end{tabular}}
 
\usepackage{subfigure}
\usepackage{multirow}
\usepackage{hyperref}
\usepackage{xspace}
\usepackage{booktabs} 
\usepackage{multicol}

\newcommand{\patks}[1]{\textsc{pass}@$#1$\xspace}
\newcommand{\ekt}[0]{EKT\xspace}
\newcommand{\gsm}{GSM8k\xspace}

\iffalse
    \newcommand{\an}[1]{\textcolor{cyan}{}}
    \newcommand{\za}[1]{\textcolor{green}{}}
    \newcommand{\hs}[1]{\textcolor{orange}{}}
\else
    \newcommand{\an}[1]{\textcolor{cyan}{\bf\small [AN: #1]}}
    \newcommand{\za}[1]{\textcolor{violet}{\bf\small [ZA: #1]}}
    \newcommand{\hs}[1]{\textcolor{orange}{\bf\small [HS: #1]}}
\fi

\iclrfinalcopy % Uncomment for camera-ready version, but NOT for submission.
\begin{document}
\maketitle
\begin{abstract}
Large language models (LLMs) can acquire strong code-generation capabilities through few-shot learning. In contrast, supervised fine-tuning is still needed for smaller models to achieve good performance and such fine-tuning demands a large number of task-specific NL-code pairs, which are expensive to obtain. In this paper, we attempt to transfer the code generation ability of an LLM to a smaller model with the aid of weakly-supervised data. More specifically, we propose {\it explicit knowledge transfer} (\ekt), which uses the few-shot capabilities of a teacher LLM to create NL-code pairs that we then filter for correctness and fine-tune the student on. We evaluate \ekt on the task of generating code solutions to math word problems from the GSM8k dataset. We find that \ekt not only yields better performance than training with expert iteration, but also outperforms knowledge distillation, another form of knowledge transfer. A GPT-Neo 1.3B model trained using \ekt with a GPT-J teacher achieves a 12.4\% \patks{100} on GSM8k, while the same student and teacher trained with knowledge distillation yield only a 3.7\% \patks{100}. We also show that it is possible for a student model to outperform the teacher using \ekt. 
% For example, a GPT-Neo 1.3B model trained with \ekt using a GPT-J teacher achieves a 1.8\% \patks{1}, while the GPT-J teacher only achieves a 0.8\% \patks{1}. 
\end{abstract}

\section{Introduction}
%\an{General comment for the intro, we want to answer four questions: 1) what is the problem we are trying to solve; 2) what are the existing methods are why they doesn't work well; 3) what is the method we are proposing and why it targets the problems that previous methods have; 4) what experiments did we do to show that our proposed method works.}
Code generation is the task of solving problems described in natural language (NL) by generating and subsequently executing code solutions written in a general-purpose programming language. 
Pretrained language models have demonstrated impressive code generation capabilities within two different paradigms. In {\it few-shot learning}, a model is conditioned to generated code by inserting a small number of NL-code pairs in its context. This method enables flexible generalization to new tasks, but only performs well with large language models (LLMs) that are computationally expensive at inference \citep{brown2020language, wei2022emergent}. The other paradigm, {\it supervised fine-tuning}, can achieve strong performance with smaller models. However, fine-tuning requires large amount of labelled training data, which is costly to manually collect and annotate \citep{chen, xu2022systematic}. 

%One popular way to scale up the training data collection is to create datasets with only 
We look to training with weak supervision as a way to achieve strong performance with small models whilst alleviating the data collection needs of supervised fine-tuning. Training with weak supervision means that only the natural language input and the expected execution result are provided for learning, leaving the gold programs latent \citep{pasupat2015compositional}. Though such weakly-supervised training data is cheaper to obtain, the learning signal from it can be weak and noisy \citep{ni2020merging}. 

In this paper, we propose {\it Explicit Knowledge Transfer} (\ekt), a method that trains a language model on weakly-supervised data with the aid of a black-box teacher LLM. In particular, \ekt uses the few-shot capabilities of an LLM to create NL-code pairs that we then filter for correctness and fine-tune on. \ekt retains the compute and memory advantages of small models at inference while avoiding the need to create large fully-labelled datasets. 
We compare \ekt to knowledge-distillation \citep{hinton, sanh}, an alternative method for leveraging a teacher model that trains the output distribution of the student against the output distribution of the teacher.  Additionally, we evaluate \ekt against expert iteration \citep{silver, polu}, which is a method that directly learns from weak supervision.
% We use Grade-school-math (GSM8k) \citep{gsm8k} as the benchmark for evaluation, where we generate Python program solutions to solve math problems described in natural language. 

Results on the Grade-school-math (GSM8k) \citep{gsm8k} dataset show that \ekt not only improves the performance of a student model learned with weak supervision, but also yields better performance than knowledge distillation from the same teacher LLM. By transferring from a GPT-J 6B model \citep{gpt-j}, \ekt is able to boost the performance of a GPT-Neo 1.3B \citep{gpt-neo} student from 2\% pass@100 to 12.4\%, while knowledge distillation only marginally improves it to 3.7\%. We also perform an ablation study for \ekt by varying the model sizes for both the student and the teacher. Our best performance is achieved with \ekt by transferring from the Codex model \citep{chen} to GPT-Neo 1.3B, which boosts the performance to 32.9\% pass@100.
% In this paper, we target the problem of generating code solutions to models to GSM8k \citep{gsm8k}, a dataset of grade-school math word problems. Because the training examples of GSM8k do not contain code solutions, this is a case where weakly-supervised learning is necessary for small models to achieve strong performance. 
% We find that on the GSM8k dataset, 
% \ekt greatly outperforms methods that do not leverage knowledge transfer. For example, a GPT-Neo 1.3B model is able to achieve up to a 18.7\% accuracy when trained with an OpenAI Codex teacher, while few-shot learning and expert iteration achieve $1.4\%$ and $0\%$ accuracy, respectively. Moreover, \ekt beats knowledge distillation in all our experiments. For example, a GPT-Neo 1.3B model with a GPT-J teacher yields an accuracy of $0.1\%$, while training with \ekt yields an accuracy of $1.8\%$. 
\begin{figure}[t]
\centering
\includegraphics[width=8cm]{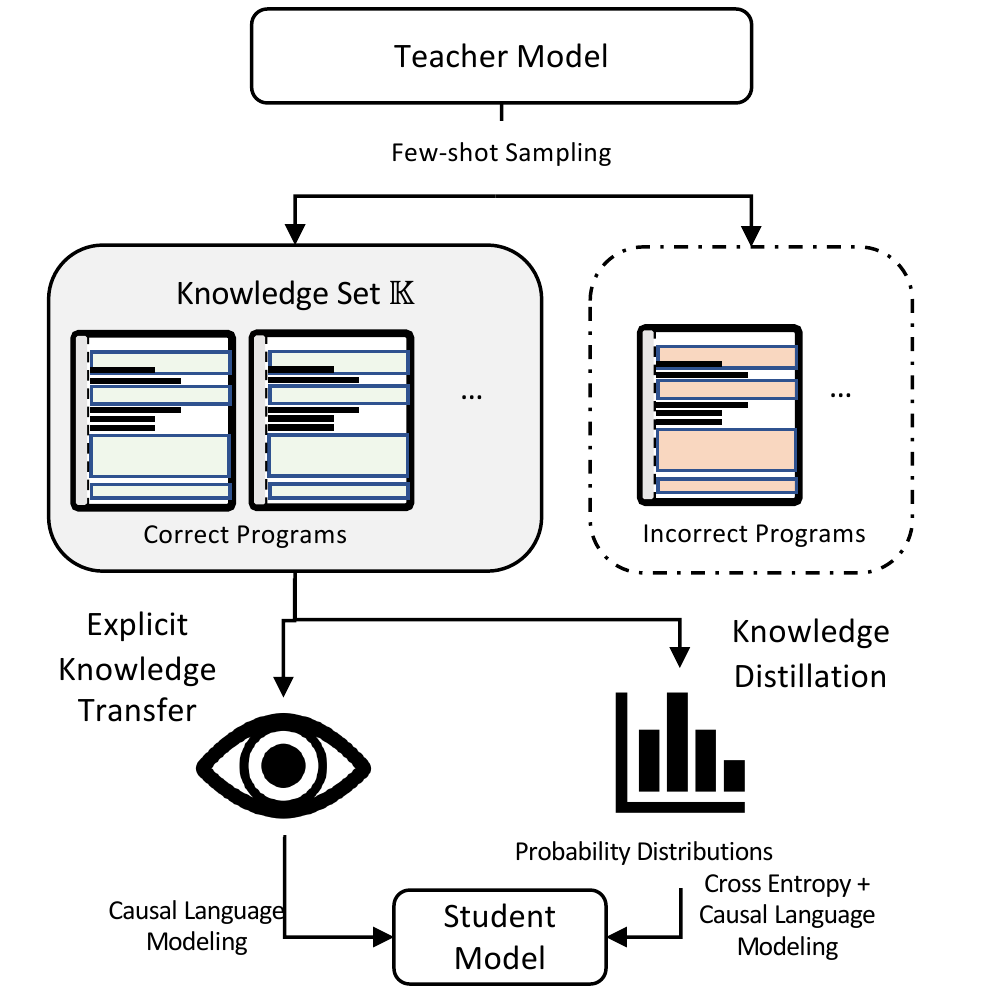}
\centering
\caption{Schematic depiction of knowledge distillation (right) and explicit knowledge transfer (left) for weakly-supervised code generation.}
\end{figure}

\section{Methodology}
We describe the weakly-supervised code generation task in \autoref{sec:formulation} and existing methods for tackling the problem in \autoref{sec:existing-methods}. We introduce knowledge transfer methods, including \ekt in \autoref{sec:kt}.
\subsection{Problem Formulation}
\label{sec:formulation}
%\an{Here you need to clearly define the *weakly-supervised* code generation task. Use my AAAI paper as a reference.}
Let $x$ be an NL specification and $y^*$ be the gold program that satisfies this specification by achieving the desired execution result $z$. 
We also assume access to a known, task-specific boolean function $f(y, z)$ that verifies the the correctness of a candidate program $y$ against $z$ (\textit{e.g.,} test cases). 
Weakly-supervised code generation is the task of learning a parameterized model $P(y|x; \theta)$ from $(x, z)$ pairs, leaving $y^*$ latent. In practice, we seed our weakly-supervised learner at the beginning with a small set of NL-code examples $S = \{(x_1, y_1), \dots\}$ to ensure nontrivial gradient updates.
% NL specification of a program, and let  NL specification. , and let $f(y, z)$, be a known, task-specific function that checks , outputting \texttt{True} if $y$ is correct as assessed by $z$ and \texttt{False} otherwise. 

\subsection{Existing Methods}
\label{sec:existing-methods}
\paragraph{Expert Iteration.} 
Since $P(y|x; \theta)$ cannot be directly optimized, much work in weakly-supervised learning has relied on \textit{expert iteration}  \citep{silver, liang, polu}. In expert iteration, a model is seeded with an initial set of NL-code pairs $D_0$. Training proceeds by alternating two steps: 1)  a sampling step, where we sample candidate solutions to weakly-supervised training examples from the model, then filter for correctness and 2) a training step, where the model is trained with MLE on all known correct NL-code pairs. 

\paragraph{{}Few-shot Learning.} Another alternative when few NL-code pairs are available is to ignore weakly-supervised data altogether, and do few-shot learning by formatting the examples in $S$ as a prompt. This method typically does not achieve strong performance with small models \citep{austin, wei2022emergent}. 

\subsection{Knowledge Transfer}
\label{sec:kt}
In expert iteration, the same model both generates novel correct programs and learns from these correct programs. We observe that generating novel samples is the step that bottlenecks the performance of such methods, which motivates offloading generation to a teacher LLM with few-shot learning abilities. In this section, we describe how to transfer code generation ability from a teacher model $P(y|x, \theta_t)$ to a student model $P(y|x, \theta_s)$.

\noindent{\bf Knowledge Set Acquisition.} %\subsection{Few-shot Solution Sampling and Filtering}
% \an{Maybe "Knowledge set acquisition"?}
% \an{Maybe "Explicit Knowledge set acquisition"?}\za{that might be a little confusing, since the knowledge set is also use for kd}
The following describes a procedure for creating the {\it knowledge set} $\mathbb{K}$ of a teacher LLM from a weakly-supervised dataset $\mathcal{D}=\{(x_i,z_i)\}_{i=1}^{|\mathcal{D}|}$ and some few-shot examples $S=\{(x_i, y_i\}_{i=1}^{|S|}$, where $|S|$ is small.
For each $(x, z)\in\mathcal{D}$, perform few-shot learning with $S$ and sample code solutions $C=\{\hat{y}_1,...\}$ from the teacher model, \textit{i.e.,} $\hat{y}_i \sim P(y|x, S; \theta_{t})$. Then choose one $\hat{y}^*\in C$ such that $f(\hat{y}^*, z)=1$, meaning $\hat{y}^*$ is a solution for $x$, if such a $\hat{y}^*$ exists.
\footnote{If a teacher generates multiple programs $\hat{y}$ passing the test cases $z$ for a training example $x$, we do not use any heuristics to decide which program is selected as $\hat{y}*$, and simply choose one uniformly at random.} 
Collect each $(x,\hat{y}^*, z)$ into a {\it knowledge set} $\mathbb{K}$.

\noindent{\bf Explicit Knowledge Transfer.}\label{ekt} %\subsection{Explicit Knowledge Transfer}\label{ekt}
Given a knowledge set $\mathbb{K}$ from the teacher model, \ekt with a weakly-supervised dataset $\{(x,z)\}$ proceeds as follows. 
% First, design the few-shot prompt $S$, and use a teacher LLM to create the knowledge set $\mathbb{K}$ from $\{(x,z)\}$ and $S$. 
For each NL-code pair $(x, y, z)$ in $\mathbb{K}$, we concatenate $x$ and $\hat{y}$ into a single sequence of tokens $t=(t_1, \dots, t_n)$ and train the student model $P(y|x; \theta_s)$ using a causal language modeling (CLM) objective, following recent work on code generation \citep{chen, austin, ni2022}: 
\begin{align*}
    \mathcal{L}_{\mathrm{MLE}}(t, \theta_s) &= 
    \sum_{k=1}^n H(P(t_k|t_{<k}; \theta_s), e_{t_k}) \\
    &= - \sum_{k=1}^n \log P(t_k|t_{<k}; \theta_s) 
\end{align*}
% \[\mathcal{L}_{\mathrm{MLE}}(x, y, \theta_s) = - \sum_{k=1}^n
% \log P(y_k|x, y_{<k}; \theta_s)\]
% \left(P_s\left(t_k|t_1^{k-1}\right), e_{t_k}\right)\]
Here $H$ denotes cross entropy and $e_{t_k}$ is the one-hot encoding of token $y_k$. 

\noindent{\bf Knowledge Distillation.} % \subsection{Knowledge Distillation.}
Knowledge distillation refers to a broad class of knowledge transfer methods whose loss incorporates the cross entropy between the student's output distribution and a teacher' output distribution. Knowledge distillation has emerged as one of the standard methods for knowledge transfer, as the non-modal probability masses of $P(y|x; \theta_t)$ are thought to encode rich information about the generalization power of the teacher \citep{hinton, papernot, sanh}

We compare \ekt to a knowledge distillation baseline trained on the same knowledge set $\mathbb{K}$. In particular, define the {\it distillation loss}:
\begin{align*}
    \mathcal{L}_\mathrm{CE}(t, \theta_t, \theta_s ) = \sum_{k=1}^n H(P(t_k|t_{<k}; \theta_s), P(t_k|t_{<k}; \theta_t))
\end{align*}
Following \citet{sanh}, we train our knowledge distillation student $\theta_s$ on a loss of the form $\mathcal{L} = \alpha \mathcal{L}_{\mathrm{CE}} + (1-\alpha)\mathcal{L}_{\mathrm{MLE}}$ where $0<\alpha<1$. 

Note that knowledge distillation is limited to the case where the teacher and the student share a vocabulary, while \ekt is not. 

\section{Experiments}
\subsection{Setup}
\iffalse
\label{expsetup}
\subsection{GSM8k dataset}
\begin{figure}[t]
\centering
\includegraphics[width=8cm]{images/gsm8k_ex.png}
\centering
\caption{Representative GSM8k instance paired with a representative generated code solution.}
\label{instance}
\end{figure}
\fi

{\bf Dataset.} We conduct experiments on the \gsm dataset \citep{gsm8k}, which consists of 7.4K training examples of grade-school-level math questions and their numerical answers\footnote{The original dataset also contains NL solutions but we ignore them in our setting since they are not executable.}.
We approach this problem by generating Python code solutions and attempt to learn the model with weak supervision.
% evaluate the \ekt method on the GSM8k dataset of grade-school math world problems. This dataset consists of 7473 training examples and 1319 test examples annotated with numerical answers and natural-language solutions. In our experiments, we ignore the natural-language solutions, and attempt to generate latent code solutions. 

\noindent{\bf Models and Training.} %\subsection{Models and Training}
We choose GPT-Neo models \citep{gpt-neo} of sizes 125M and 1.3B as student models, with GPT-Neo 2.7B, GPT-J 6B \citep{gpt-j} and Codex\footnote{More specifically, we the use \texttt{code-davinci-002} engine.} \citep{chen} as ablations for different teacher models.
In our experiments, we evaluate four learning methods: 1) few-shot learning; 2) expert iteration (EI); 3) knowledge distillation (KD); and 4) explicit knowledge transfer (\ekt). 
For EI, we initialize $D_0$ by using a few-shot prompt to sample from our not yet fine-tuned model $P(y|x, S; \theta)$, followed by correctness filtering. 
% \an{I would suggest cut the next two sentences} This method of initializing $D_0$ is rather limiting, as the small models we seek to train with expert iteration have minimal few-shot learning capabilities. This motivates the use of knowledge transfer. 
% We evaluate GPT-Neo 125M and GPT-Neo 1.3B \citep{gpt-neo} on GSM8k trained using each of our methods. For \ekt experiments, we train separate models using OpenAI's \texttt{code-davinci-002} engine (hereafter ``Codex''), GPT-J \citep{gpt-j}, and GPT-Neo 2.7B \citep{gpt-neo} as teachers. 

% \subsection{Evaluation Methods}
% \an{Move this to the experiment setup}\hs{I did this. can be condensed with next paragraph}
% We evaluate code generation performance using the pass@$k$ metric \citep{chen}. Given $k$ sampled programs for a given NL specification, pass@$k=1$ if at least one of the $k$ sampled programs are correct as judged by $f(y,z)$, and $0$ otherwise. To estimate pass@$k$, we use the unbiased estimator of \citet{chen}. 

\noindent{\bf Evaluation Metric.} % \subsection{Teacher Sampling and Evaluation}
% For few-shot prompting, use a 3-shot prompt that can be found in \autoref{prompts}.
% \an{Put the prev sentence in the method part.}\za{does it make sense to introduce the specific few shot prompt we're using before we talk about the dataset?}
We evaluate code generation performance using the \patks{k} metric \citep{chen}. Given $k$ sampled programs for a given NL specification, \patks{k}$=1$ if any of the programs $\hat{y}$ is correct (\textit{i.e.,} $f(\hat{y}, z)=\texttt{True}$), then the average \patks{k} is reported.

More details for the experimental setups (\textit{e.g.,} hyperparameters, few-shot prompts) can be found in \autoref{params}. 
%\subsection{Limitations of Knowledge Distillation}\label{kdlimitations}
%\an{Why is this in the setup and not the discussion?} \hs{I think that the reason for this was just to justify the lack of a Codex knowledge distillation experiment. But could be moved if that doesn't come across}
%Knowledge distillation, when applied to a causal language model, depends on the teacher and the student model sharing a vocabulary so that output logits of each token are directly comparable across models. In practice, this can be a significant limitation, as the largest state-of-the-art models likely to be used as a teacher and smaller open-sourced models likely to be used as a student may come from different model families and thus not share a vocabulary. \ekt treats the teacher as a text-in text-out black-box, and thus does not share this limitation. For this reason, we are unable to train GPT-Neo models with {\it code-davinci-002} as the teacher. 

\subsection{Knowledge Transfer Results}
%old: The results for all experiments described in section \autoref{expsetup} can be found in table \autoref{results}. It can be seen that \ekt outperforms the few-shot, expert iteration, and knowledge-distillation baselines in all experiments. Additionally, student models trained with \ekt (and sometimes even those trained with knowledge distillation) are capable of outperforming their teacher, as can be seen in the experiments with GPT-Neo 2.7B and GPT-J teachers.
\iffalse
 \begin{figure}[ht]
\centering
\begin{subfigure}(a)
   \includegraphics[width=1\linewidth]{images/pass_1.png}
   \label{fig:pass1} 
\end{subfigure}

\begin{subfigure}(b)
   \includegraphics[width=1\linewidth]{images/pass_100.png}
   \label{fig:pass100}
\end{subfigure}

\caption{Plots demonstrating the effect of student and teacher model size}
\end{figure}
\fi

\iffalse
\begin{table*}[ht]
\centering
\begin{tabular}{lcccc}
Training Method & \multicolumn{2}{c}{GPT-Neo 125M} & \multicolumn{2}{c}{GPT-Neo 1.3B}\\
&\patks{1}& \patks{100} & \patks{1} & \patks{100}\\
\hline\hline
Baselines: &&&&\\
\quad Few-shot&0.2\%&0.4\%&1.4\%&2\%\\
\quad Expert iteration &0\%&0\%&0\%&0\%\\
\hline
Knowledge Distillation: &&&&\\
\quad w/ GPT-J teacher &0\%&3.6\%&0.1\%&3.7\%\\
\hline
\ekt: &&&&\\
\quad w/ GPT-J teacher &1.1\%&13.6\%&1.8\%&12.4\%\\
\quad w/ Codex teacher &{\bf5.6\%}&{\bf18.2\%}&{\bf18.7\%}&{\bf32.9\%}\\
\hline
\end{tabular}
\caption{Comparison of \ekt to baseline methods on GSM8k.}
\label{baselines}
\end{table*}
\fi 

\begin{table}[t!]
\centering
\small
\begin{tabular}{lrr}
\toprule
\textbf{Model} & \textbf{\patks{1}} & \textbf{\patks{100}}\\
\midrule
\textit{Few-shot learning:} &&\\
\quad Minerva 540B $^{a*}$ & 58.8 & - \\
\quad PaLM 540B$^{b*}$ & 56.5 & -\\
\quad LaMDA 137B$^{c*}$ & 7.6 & -\\
\b\quad Codex-Davinci & 37.4 & 92.3\\
\quad Codex-Cushman & 5.0 & 58.0\\
\quad GPT-Neo 1.3B & 1.4 & 2.0\\
\midrule
\textit{Fine-tuned w/ full supervision:} &&\\
\quad GPT-3 6B + verifier$^{d*}$ & 39.0 &-\\
\quad GPT-3 175B + verifier$^{d*}$ & 55.0 & -\\
\quad GPT-Neo 2.7B + SS$^e$ & 19.5 & 41.4\\
\midrule
\multicolumn{3}{l}{\textit{Weakly-supervised methods (all w/ GPT-Neo 1.3B):}} \\
\quad Expert Iteration & 0.0 & 0.0 \\
% \hline
% Weak Supervision + Teacher&&\\
\quad KD from GPT-J 6B &0.1 &3.7\\
\quad \ekt from GPT-J 6B & 1.8 & 12.4\\
\quad \ekt from Codex-davinci & {\bf18.2} & {\bf32.9} \\\bottomrule
\end{tabular}
\caption{Comparison of \ekt to baseline methods on test set of \gsm. Note that we are unable to evaluate KD from Codex-davinci to GPT-Neo, since the models have different vocabularies. $^a$: \citet{minerva} $^b$: \citet{palm}; $^c$: \citet{lamda}; $^d$: \citet{gsm8k}; $^e$: SS denotes self-sampling from \citet{ni2022}; $^*$: models that generate NL solutions instead of code; -: no results are available.}
\label{baselines}
\end{table}

% \begin{table*}[ht]
% \centering
% \begin{tabular}{lccc|cc}
% &\multicolumn{3}{c}{Teacher} & \multicolumn{2}{c}{Student}  \\
% Models & Training set coverage & \patks{1} & \patks{100} & \patks{1} & \patks{100}\\
% \hline\hline
% GPT-Neo 2.7B teacher &19.5\%&0.4\%&13\%&&\\
% \quad w/ GPT-Neo 125M student&&&&0.3\%&10.8\%\\
% \quad w/ GPT-Neo 1.3B student&&&&1.1\%&8.7\%\\
% \hline
% GPT-J teacher &39.7\%&0.8\%&&&\\
% \quad w/ GPT-Neo 125M student&&&&1.1\%&13.6\%\\
% \quad w/ GPT-Neo 1.3B student&&&&1.8\%&12.4\%\\
% \hline
% {\it code-davinci-002} teacher &93.0\%&37.4\%&92.3\%&&\\
% \quad w/ GPT-Neo 125M student&&&&5.6\%&18.2\%\\
% \quad w/ GPT-Neo 1.3B student&&&&18.7\%&32.9\%\\
% \end{tabular}

% \caption{\ekt results with all teacher and student pairs\an{maybe it's possible to replace this with a figure? e.g., x axis being the model size and y axis being the \patks{k} numbers, and 125M/1.3B being two separate lines?} \za{I like this idea, the only issue is that there may be too many numbers in the <2\% range for a graph to be viable}}
% \label{ablation}
% \end{table*}
\begin{table}[t]
\centering
\small
\begin{tabular}{lrrr}
\toprule
\textbf{\textcolor{gray}{Teacher model}} & \multirow{2}{*}{\textbf{\textcolor{gray}{Coverage}}} & \multirow{2}{*}{\textbf{\patks{1}}} & \multirow{2}{*}{\textbf{\patks{100}}}\\
\quad \text{Student model} &&&\\
\midrule
\textcolor{gray}{GPT-Neo 2.7B} & \textcolor{gray}{19.5} & \textcolor{gray}{0.4} & \textcolor{gray}{13.0}\\
\quad GPT-Neo-125M & - & 0.3 &10.8 \\
\quad GPT-Neo-1.3B & - & 1.1 & 8.7\\
\midrule
\textcolor{gray}{GPT-J 6B} & \textcolor{gray}{39.7} & \textcolor{gray}{0.8} & \textcolor{gray}{30.6}\\
\quad GPT-Neo-125M & - & 1.1 & 13.6 \\
\quad GPT-Neo-1.3B & - & 1.8 & 12.4\\
\midrule
\textcolor{gray}{Codex-Davinci} & \textcolor{gray}{93.0} & \textcolor{gray}{37.4} & \textcolor{gray}{92.3}\\
\quad GPT-Neo-125M & - & 5.6 &18.2 \\
\quad GPT-Neo-1.3B & - & 18.7 & 32.9\\
\bottomrule
\end{tabular}\caption{Ablation study demonstrating the effect of modulating the teacher and the student when training with \ekt. {\it Coverage} is the percentage of training examples where a teacher generated a correct program.}
\label{ablation}
\end{table}

{\bf Effectiveness of EKT.} In \autoref{baselines}, we compare to performance of \ekt to various baselines. \ekt outperforms all other weakly-supervised training methods. We offer a hypothesis as to why \ekt outperforms knowledge distillation despite learning from the same knowledge set. Because we sample from the teacher at a high temperature to generate the knowledge set, it may be that the teacher generates a correct program by sampling from a region of low probability mass. Since \ekt trains with MLE, the student will ``disagree'' with the teacher and assign this generated program a high probability. In contrast, knowledge distillation will mimic the output distribution of the teacher, and thus assign a low probability to all programs the teacher assigned a low probability to. 

Despite the strong performance of expert iteration in other settings, %code generation and completion tasks \citep{liang, polu}
it does not achieve nontrivial performance in our experiments. The poor performance of expert iteration is explained by the fact that the seed dataset $D_0$ is too small for the model to achieve non-trivial bootstrapping ability. For example, in expert iteration with GPT-Neo 1.3B, we see that $|D_0|=|D_1|=43$, suggesting the model is unable to generalize when fine-tuning on only 43 examples.

\subsection{Ablation on Student and Teacher Models}
In \autoref{ablation} we present an ablation study that demonstrates the effect of modulating the size of the teacher and of the student. We observe that when using \ekt, increasing the strength of the teacher is more effective than increasing the capacity of the student. 
For example, using a GPT-Neo 125M model with a GPT-J 6B teacher as a baseline, switching to a GPT-Neo 1.3B student only improves \patks{1} by 0.7\% while upgrading to a Codex teacher increases \patks{1} by 4.4\% and \patks{100} from 12.4\% to 18.2\%. We believe this is because more capable teacher provides better coverage (\textit{i.e.,} 39.7\% and 93.0\% in this case), which enables the student to learn from more training examples.
The hypothesis is supported by the fact the 125 million parameter student has a \patks{1} close to the 1.3 billion parameter student and a slightly higher \patks{100} when transferring from a GPT-J teacher using \ekt. In this case, increasing the capacity of the student in a low-data regime may lead to a degraded \patks{100} because a stronger model overfits more easily and thus does not preserve sample diversity.

% for some less performing teachers, upgrading the capacity of the student may not increase performance.
% In contrast, increasing the parameter count of the student by roughly a factor of 10 through upgrading to a GPT-Neo 1.3B student only increases \patks{1} by only 1.2 percentage points. 
% Additionally, for some less performing teachers, upgrading the capacity of the student may not increase performance. For example, with \ekt using a GPT-J teacher, the 125 million parameter student has a \patks{1} close to the 1.3 billion parameter student and a slightly higher \patks{100}. We hypothesize that increasing the capacity of the student in a low-data regime leads to a degraded \patks{100} because the stronger model overfits more easily and thus does not preserve sample diversity.

In the case of GPT-Neo 2.7B and GPT-J, both student models outperform their \ekt teacher at \patks{1}. This is not unexpected for two reasons. First, the student model learns from correct solutions generated by the teacher in the setting of a high sampling budget and a high sampling temperature. This means that the student can learn from programs that the teacher is unlikely to generate in the setting of low-temperature or greedy decoding. 
Second, the student is conditioned on more task specific data, namely the teacher's knowledge set. 

\iffalse
{\bf Code Solution Analysis} 
\autoref{fig:teacher_lens} depicts the length distribution of programs discovered by a Codex teacher and its student on the test set, along with those generated by a GPT-Neo 2.7B teacher on the train set (for better estimation of a low-powered teacher's distribution) and student on the test set. We normalize the 2.7B model's Y axis to match its student, since the train set has roughly 5 times more problems to solve than the test set. 

We observe that the length distribution of GPT-Neo 2.7B solutions on the train set peaks close 7 and 20 lines, which correspond to the lengths of solutions in our few shot prompt. In contrast, the Codex teacher's solutions have a smoother distribution that likely better approximates the true latent distribution of solution lengths. Both of these teacher's length distributions are reflected by their respective student. This gives an additional hint as to why a better teacher greatly improves student performance: strong teachers do not just provide the student with more data, but also better approximate the latent data-generating distribution.

 \begin{figure}[h]
 \centering
 \includegraphics[width=\linewidth]{images/student_teacher.png}
 \includegraphics[width=\linewidth]{images/neo_student_teacher_weighted.png}
 \caption{Distribution of Found Solution Complexities}
 \label{fig:teacher_lens}
 \centering
 \end{figure}
\fi
\section{Related Work}
\paragraph{Language models trained on code.} Code LMs have achieved strong code generation performance with few-shot learning in the case of LLMs and supervised fine-tuning in the case of small models \citep{chen, austin, li, nijkamp, fried}.

\paragraph{Weakly-supervised semantic parsing.} Semantic parsing is the task of mapping an NL utterance to an executable formal representation, possibly not in a general purpose language, and the weakly-supervised case is a well-studied problem in NLP  \citep{zelle, zettlemoyer, zhong, liang, dong}.

%  Expert iteration is a method of weakly-supervised learning where a model bootstraps itself by learning from its previous successful samples. This training method was introduced in \citet{silver} for playing board games, but variants of the methodology have been applied to code generation in Lisp \citep{liang}, interactive theorem proving \citep{polu}, and NL reasoning \citep{wu}. %In expert iteration, the same model both generates novel successful samples and learns from previous successful samples. We observe that generating novel successful samples is often much more difficult than learning from previous successful samples, therefore we task a stronger model with generation and a weaker model with doing supervised learning on the stronger model's generations. 

\paragraph{Knowledge Transfer} 
% Hailey comment: does this paragraph focus too much on what we do rather than related work?
{\it Knowledge distillation} is a popular method of knowledge transfer for neural networks \citep{hinton, papernot, sanh}. %alternative choices for distillation loss aside from logits
%The most well-known application of knowledge distillation applied to transformer-based language models is DistilBERT \citet{sanh}, in which a teacher logits-matching loss described in \citep{hinton} % insert equation here is used at the pre-training stage to train a model with 40\% fewer parameters but retaining 97\% of the downstream performance of the teacher model. This method attempts to train a student model by using this loss term to push the output logits of the student to be close to the teacher, thus theoretically resulting in a student with similar behavior (and performance) to a teacher model. However, the method used in DistilBERT relies on an initialization of the layers of the student using selected layers from the teacher, and thus prevents a large, multi-billion parameter model likely to achieve strong few-shot performance % cite emergence paper being distilled into a student model with less than a billion parameters suitable for fine-tuning on a low-resource compute setup.

\section{Conclusion}
We proposed \ekt, a method of weakly-supervised code generation that leverages a teacher LLM. We showed that \ekt outperforms weakly-supervised baselines and other knowledge transfer methods, with a GPT-Neo 1.3B student trained using \ekt with a Codex-davinci teacher achieving a \patks{1} of 18.2\% on \gsm. Additionally, we studied the effects of modulating the size of the student and the teacher model. 

\section*{Acknowledgements} We would like to thank OpenAI for providing access to the Codex API public beta.

\iffalse 
\section{Limitations}
One limitation of \ekt is the risk of including solutions that arrive at the right answer by incorrect reasoning in the knowledge set. Because answers to \gsm problems are floating point numbers, we do not have strong reason to suspect spurious solutions are a serious problem in our experiments, since it is exceedingly unlikely for an arbitrary sequence of arithmetic operations to yield a given float. However, if one is applying \ekt to programs with discrete output, this may be a significant problem. 
\fi 
% \section*{Acknowledgements}
%\an{You don't do acknowledgements in the anonymous review version.}
\bibliography{iclr2023_conference}
\bibliographystyle{iclr2023_conference}

\appendix

\begin{table}[h]
\fontfamily{qcr}\selectfont
\tiny
\begin{tabular}{|p{\linewidth}|}
\hline\\
\# The total average age of three friends is 40. Jared is ten years older than Hakimi, and Molly's age is 30. How old is Hakimi?\\
n0 = 40 \\
n1 = 10 \\
n2 = 30 \\
t0 = 3 * n0\\ 
t1 = t0 - n2\\
answer = (t1 - n1) / 2\\\\

\# A carpenter worked alone for 1 day on a job that would take him 7 more days to finish. He and another carpenter completed the job in 4 more days. How many days would it have taken the second carpenter to do the complete job working alone?\\ 
n0 = 1.0\\
n1 = 7.0\\
n2 = 4.0\\
t0 = n0 + n1\\
t1 = n2 * t0\\
answer = t1 / 2.0\\\\

\# In two alloys, copper and tin are related in the ratios of 4 : 1 and 1 : 3. 10 kg of 1st alloy, 16 kg of the 2nd alloy and some pure copper are melted together. An alloy is obtained in which the ratio of copper and tin was 3 : 2 . Find the weight of the new alloy. \\
n0 = 4.0\\
n1 = 1.0\\
n2 = 1.0\\
n3 = 3.0\\
n4 = 10.0\\
n5 = 16.0\\
n6 = 2.0\\
n7 = 3.0\\
n8 = 2.0\\
t0 = n4 + n5\\
t1 = n0 + n1\\
t2 = n3 / n0\\
t3 = n4 / t1\\
t4 = n5 * t2\\
t5 = t3 + t4\\
t6 = n3 * t5\\
t7 = t6 / n6\\
t8 = t7 - t4\\
answer = t0 + t8\\\\
\hline
\end{tabular}
\small\fontfamily{qtm}\selectfont
\caption{\small\fontfamily{qtm}\selectfont Our few-shot prompt $S$.}
\label{prompt}
\end{table}

\section{Hyperparameters}
\label{params}
% placeholder: \patks{k} standardization: 
%Teachers: \patks{100} temp 0.6, 
%Students: 
%Validation: \patks{10} temp 0.2 
%Training hyperparameters: see appendix a
%Retrain on train+dev if time (highly optional). 
\subsection{Training}
We use the hyperparameters in \autoref{tabparams} to train our student models, both in the \ekt and knowledge distillation case. 
\begin{table*}[h]
\centering
\begin{tabular}{c|c}
Parameter & Setting\\
\hline 
Training Epochs & 140\\
Learning Rate (LR) & $1\cdot 10^{-4}$\\
Optimizer & AdamW\\
Adam Betas & $(0.9, 0.999)$\\
Adam Eps & $1\cdot 10^{-8}$\\
Weight Decay & $0.1$\\
LR Scheduler & Linear w/ warm-up\\
LR Warm-up Steps & 100\\
Effective Batch Size & 32\\
Precision & FP32\\
Gradient Clipping & 1.0\\
\end{tabular}
\caption{Student training hyperparameters.}
\label{tabparams}
\end{table*}

For expert iteration, each $M_n$ is trained using the same hyperparameters as in table $\autoref{tabparams}$, except we train for 4 epochs and use a fixed learning rate of $5\cdot 10^{-5}$. We stop expert iteration at the iteration $N$ where $|\mathbb{K}_{N-1}|=|\mathbb{K}_N|$. After stopping iteration, we further train $M_N$ on $\mathbb{K}_N$ using the hyperparameters in table \autoref{tabparams} to yield our final trained model. 

For knowledge distillation, we use a weighting factor of $\alpha=1/2$. 

For our experiments, we reserve 500 randomly-sampled training examples as a validation set, and so do not train our models on those examples. 

\subsection{Sampling and Evaluation}
We use the few shot prompt in table \autoref{prompt}. To generate the knowledge set $\mathbb{K}$ from our teacher LLMs, we generate 100 samples per training example using temperature sampling at a temperature of 0.6 \citep{ackley, ficler}. To calculate \patks{1} on the test set, we generate samples with greedy decoding and for \patks{100} on the test set, we generate samples using temperature sampling with a temperature of 0.6. 

For expert iteration, we create the initial knowledge-set $D_0$ by sampling from our not yet fine-tuned model $P(y|x, S; \theta)$ with 100 samples per question and a temperature of 0.6 on the \gsm training set and filtering for correctness.

\section{Complete Results}
For complete results of all our experiments, see \autoref{tab:results}. 
\label{complete_results}
\begin{table*}[t]
\small
\centering
\begin{tabular}{|c|c|c|ccccc|}
\hline
\textbf{\multirow{2}{*}{\specialcell{Teacher\\model}}}  & \textbf{\multirow{2}{*}{\specialcell{Student\\model}}} & 
\textbf{\multirow{2}{*}{\specialcell{Student\\training\\method}}} &  \multicolumn{3}{|c|}{Teacher} &\multicolumn{2}{|c|}{Student} \\
\cline{4-8} 
& &  &  {\bf \specialcell{training set\\coverage}} & {\bf \patks{1}} & {\bf \patks{100}} &{\bf \patks{1}}& {\bf \patks{100}}\\
\hline 
\hline
{\it None} &  \multirow{7}{*}{\specialcell{GPT-Neo\\125M}} & Few-shot &- &-&-  &0.2\%&0.4\%\\
{\it None}: &   & EI & -&-&-&& \\
GPT-Neo 2.7B &   & KD & 19.5\%&0.4\%& 13\%& 0\% &2\%\\
GPT-Neo 2.7B &    & \ekt & 19.5\%&0.4\%&13\%&0.3\%&10.8\%\\
GPT-J &   & KD & 39.7\%&0.8\%&30.6\% & 0\% &3.6\%\\
GPT-J &    & \ekt & 39.7\%&0.8\%&30.6\%&1.1\%&13.6\%\\
{\it \specialcell{Davinci-\\code-002}} &   & \ekt & 93.0\%&37.4\%&92.3\%&5.6\%& 18.2\%\\
\hline 
{\it None}  & \multirow{7}{*}{\specialcell{GPT-Neo\\1.3B}} & Few-shot & - &- &- & 1.4\%&\\
{\it None}  &  & EI & - &- &-&& \\
GPT-Neo 2.7B &   & KD &19.5\% &0.4\%&13\%& 0\% &2.8\%\\
GPT-Neo 2.7B &    & \ekt &19.5\% &0.4\%&13\% &1.1\%&8.7\%\\
GPT-J  &  & KD &39.7\% &0.8\%&30.6\%& 0.1\% & 3.7\% \\
GPT-J   &  & \ekt &39.7\%  &0.8\%&30.6\% &1.8\%&12.4\%\\
{\it \specialcell{Davinci-\\code-002}} &   & \ekt & 93.0\% &37.4\%&92.3\%&18.7\%&32.9\% \\
\hline\hline
\end{tabular}
\caption{Complete results of our experiments on the GSM8k dataset. EI: expert iteration; KD: knowledge distillation. {\it Training set coverage} denotes the percentage of training examples where the teacher generated at least one correct program during sampling.}
\label{tab:results}
\end{table*}

\end{document}

%% file: math_commands.tex
\usepackage{amsmath,amsfonts,bm}

\def\eqref#1{equation~\ref{#1}}
\def\1{\bm{1}}

\DeclareMathAlphabet{\mathsfit}{\encodingdefault}{\sfdefault}{m}{sl}
\SetMathAlphabet{\mathsfit}{bold}{\encodingdefault}{\sfdefault}{bx}{n}

%% file: iclr2023_conference.bbl
\begin{thebibliography}{29}
\providecommand{\natexlab}[1]{#1}
\providecommand{\url}[1]{\texttt{#1}}
\expandafter\ifx\csname urlstyle\endcsname\relax
  \providecommand{\doi}[1]{doi: #1}\else
  \providecommand{\doi}{doi: \begingroup \urlstyle{rm}\Url}\fi

\bibitem[Ackley et~al.(1985)Ackley, Hinton, and Sejnowski]{ackley}
David~H. Ackley, Geoffrey~E. Hinton, and Terrence~J. Sejnowski.
\newblock A learning algorithm for boltzmann machines.
\newblock \emph{Cogn. Sci.}, 9:\penalty0 147--169, 1985.

\bibitem[Austin et~al.(2021)Austin, Odena, Nye, Bosma, Michalewski, Dohan,
  Jiang, Cai, Terry, Le, and Sutton]{austin}
Jacob Austin, Augustus Odena, Maxwell Nye, Maarten Bosma, Henryk Michalewski,
  David Dohan, Ellen Jiang, Carrie~J. Cai, Michael Terry, Quoc~V. Le, and
  Charles Sutton.
\newblock Program synthesis with large language models.
\newblock \emph{CoRR}, abs/2108.07732, 2021.
\newblock URL \url{https://arxiv.org/abs/2108.07732}.

\bibitem[Black et~al.(2021)Black, Gao, Wang, Leahy, and Biderman]{gpt-neo}
Sid Black, Leo Gao, Phil Wang, Connor Leahy, and Stella Biderman.
\newblock {GPT-Neo}: Large scale autoregressive language modeling with
  mesh-tensorflow, 2021.
\newblock URL \url{http://github.com/eleutherai/gpt-neo}.

\bibitem[Brown et~al.(2020)Brown, Mann, Ryder, Subbiah, Kaplan, Dhariwal,
  Neelakantan, Shyam, Sastry, Askell, Agarwal, Herbert-Voss, Krueger, Henighan,
  Child, Ramesh, Ziegler, Wu, Winter, Hesse, Chen, Sigler, Litwin, Gray, Chess,
  Clark, Berner, McCandlish, Radford, Sutskever, and Amodei]{brown2020language}
Tom~B. Brown, Benjamin Mann, Nick Ryder, Melanie Subbiah, Jared Kaplan,
  Prafulla Dhariwal, Arvind Neelakantan, Pranav Shyam, Girish Sastry, Amanda
  Askell, Sandhini Agarwal, Ariel Herbert-Voss, Gretchen Krueger, Tom Henighan,
  Rewon Child, Aditya Ramesh, Daniel~M. Ziegler, Jeffrey Wu, Clemens Winter,
  Christopher Hesse, Mark Chen, Eric Sigler, Mateusz Litwin, Scott Gray,
  Benjamin Chess, Jack Clark, Christopher Berner, Sam McCandlish, Alec Radford,
  Ilya Sutskever, and Dario Amodei.
\newblock Language models are few-shot learners, 2020.

\bibitem[Chen et~al.(2021)Chen, Tworek, Jun, Yuan, Pinto, Kaplan, Edwards,
  Burda, Joseph, Brockman, Ray, Puri, Krueger, Petrov, Khlaaf, Sastry, Mishkin,
  Chan, Gray, Ryder, Pavlov, Power, Kaiser, Bavarian, Winter, Tillet, Such,
  Cummings, Plappert, Chantzis, Barnes, Herbert-Voss, Guss, Nichol, Paino,
  Tezak, Tang, Babuschkin, Balaji, Jain, Saunders, Hesse, Carr, Leike, Achiam,
  Misra, Morikawa, Radford, Knight, Brundage, Murati, Mayer, Welinder, McGrew,
  Amodei, McCandlish, Sutskever, and Zaremba]{chen}
Mark Chen, Jerry Tworek, Heewoo Jun, Qiming Yuan, Henrique Ponde de~Oliveira
  Pinto, Jared Kaplan, Harri Edwards, Yuri Burda, Nicholas Joseph, Greg
  Brockman, Alex Ray, Raul Puri, Gretchen Krueger, Michael Petrov, Heidy
  Khlaaf, Girish Sastry, Pamela Mishkin, Brooke Chan, Scott Gray, Nick Ryder,
  Mikhail Pavlov, Alethea Power, Lukasz Kaiser, Mohammad Bavarian, Clemens
  Winter, Philippe Tillet, Felipe~Petroski Such, Dave Cummings, Matthias
  Plappert, Fotios Chantzis, Elizabeth Barnes, Ariel Herbert-Voss,
  William~Hebgen Guss, Alex Nichol, Alex Paino, Nikolas Tezak, Jie Tang, Igor
  Babuschkin, Suchir Balaji, Shantanu Jain, William Saunders, Christopher
  Hesse, Andrew~N. Carr, Jan Leike, Josh Achiam, Vedant Misra, Evan Morikawa,
  Alec Radford, Matthew Knight, Miles Brundage, Mira Murati, Katie Mayer, Peter
  Welinder, Bob McGrew, Dario Amodei, Sam McCandlish, Ilya Sutskever, and
  Wojciech Zaremba.
\newblock Evaluating large language models trained on code, 2021.
\newblock URL \url{https://arxiv.org/abs/2107.03374}.

\bibitem[Chowdhery et~al.(2022)Chowdhery, Narang, Devlin, Bosma, Mishra,
  Roberts, Barham, Chung, Sutton, Gehrmann, Schuh, Shi, Tsvyashchenko, Maynez,
  Rao, Barnes, Tay, Shazeer, Prabhakaran, Reif, Du, Hutchinson, Pope, Bradbury,
  Austin, Isard, Gur-Ari, Yin, Duke, Levskaya, Ghemawat, Dev, Michalewski,
  Garcia, Misra, Robinson, Fedus, Zhou, Ippolito, Luan, Lim, Zoph, Spiridonov,
  Sepassi, Dohan, Agrawal, Omernick, Dai, Pillai, Pellat, Lewkowycz, Moreira,
  Child, Polozov, Lee, Zhou, Wang, Saeta, Diaz, Firat, Catasta, Wei,
  Meier-Hellstern, Eck, Dean, Petrov, and Fiedel]{palm}
Aakanksha Chowdhery, Sharan Narang, Jacob Devlin, Maarten Bosma, Gaurav Mishra,
  Adam Roberts, Paul Barham, Hyung~Won Chung, Charles Sutton, Sebastian
  Gehrmann, Parker Schuh, Kensen Shi, Sasha Tsvyashchenko, Joshua Maynez,
  Abhishek Rao, Parker Barnes, Yi~Tay, Noam Shazeer, Vinodkumar Prabhakaran,
  Emily Reif, Nan Du, Ben Hutchinson, Reiner Pope, James Bradbury, Jacob
  Austin, Michael Isard, Guy Gur-Ari, Pengcheng Yin, Toju Duke, Anselm
  Levskaya, Sanjay Ghemawat, Sunipa Dev, Henryk Michalewski, Xavier Garcia,
  Vedant Misra, Kevin Robinson, Liam Fedus, Denny Zhou, Daphne Ippolito, David
  Luan, Hyeontaek Lim, Barret Zoph, Alexander Spiridonov, Ryan Sepassi, David
  Dohan, Shivani Agrawal, Mark Omernick, Andrew~M. Dai,
  Thanumalayan~Sankaranarayana Pillai, Marie Pellat, Aitor Lewkowycz, Erica
  Moreira, Rewon Child, Oleksandr Polozov, Katherine Lee, Zongwei Zhou, Xuezhi
  Wang, Brennan Saeta, Mark Diaz, Orhan Firat, Michele Catasta, Jason Wei,
  Kathy Meier-Hellstern, Douglas Eck, Jeff Dean, Slav Petrov, and Noah Fiedel.
\newblock Palm: Scaling language modeling with pathways, 2022.
\newblock URL \url{https://arxiv.org/abs/2204.02311}.

\bibitem[Cobbe et~al.(2021)Cobbe, Kosaraju, Bavarian, Hilton, Nakano, Hesse,
  and Schulman]{gsm8k}
Karl Cobbe, Vineet Kosaraju, Mohammad Bavarian, Jacob Hilton, Reiichiro Nakano,
  Christopher Hesse, and John Schulman.
\newblock Training verifiers to solve math word problems.
\newblock \emph{CoRR}, abs/2110.14168, 2021.
\newblock URL \url{https://arxiv.org/abs/2110.14168}.

\bibitem[Dong \& Lapata(2018)Dong and Lapata]{dong}
Li~Dong and Mirella Lapata.
\newblock Coarse-to-fine decoding for neural semantic parsing.
\newblock In \emph{Proceedings of the 56th Annual Meeting of the Association
  for Computational Linguistics (Volume 1: Long Papers)}, pp.\  731--742,
  Melbourne, Australia, July 2018. Association for Computational Linguistics.
\newblock \doi{10.18653/v1/P18-1068}.
\newblock URL \url{https://aclanthology.org/P18-1068}.

\bibitem[Ficler \& Goldberg(2017)Ficler and Goldberg]{ficler}
Jessica Ficler and Yoav Goldberg.
\newblock Controlling linguistic style aspects in neural language generation,
  2017.
\newblock URL \url{https://arxiv.org/abs/1707.02633}.

\bibitem[Fried et~al.(2022)Fried, Aghajanyan, Lin, Wang, Wallace, Shi, Zhong,
  Yih, Zettlemoyer, and Lewis]{fried}
Daniel Fried, Armen Aghajanyan, Jessy Lin, Sida Wang, Eric Wallace, Freda Shi,
  Ruiqi Zhong, Wen-tau Yih, Luke Zettlemoyer, and Mike Lewis.
\newblock Incoder: A generative model for code infilling and synthesis, 2022.
\newblock URL \url{https://arxiv.org/abs/2204.05999}.

\bibitem[Hinton et~al.(2015)Hinton, Vinyals, and Dean]{hinton}
Geoffrey Hinton, Oriol Vinyals, and Jeff Dean.
\newblock Distilling the knowledge in a neural network, 2015.
\newblock URL \url{https://arxiv.org/abs/1503.02531}.

\bibitem[Lewkowycz et~al.(2022)Lewkowycz, Andreassen, Dohan, Dyer, Michalewski,
  Ramasesh, Slone, Anil, Schlag, Gutman-Solo, Wu, Neyshabur, Gur-Ari, and
  Misra]{minerva}
Aitor Lewkowycz, Anders Andreassen, David Dohan, Ethan Dyer, Henryk
  Michalewski, Vinay Ramasesh, Ambrose Slone, Cem Anil, Imanol Schlag, Theo
  Gutman-Solo, Yuhuai Wu, Behnam Neyshabur, Guy Gur-Ari, and Vedant Misra.
\newblock Solving quantitative reasoning problems with language models, 2022.
\newblock URL \url{https://arxiv.org/abs/2206.14858}.

\bibitem[Li et~al.(2022)Li, Choi, Chung, Kushman, Schrittwieser, Leblond,
  Eccles, Keeling, Gimeno, Lago, Hubert, Choy, d'Autume, Babuschkin, Chen,
  Huang, Welbl, Gowal, Cherepanov, Molloy, Mankowitz, Robson, Kohli,
  de~Freitas, Kavukcuoglu, and Vinyals]{li}
Yujia Li, David Choi, Junyoung Chung, Nate Kushman, Julian Schrittwieser, Rémi
  Leblond, Tom Eccles, James Keeling, Felix Gimeno, Agustin~Dal Lago, Thomas
  Hubert, Peter Choy, Cyprien de~Masson d'Autume, Igor Babuschkin, Xinyun Chen,
  Po-Sen Huang, Johannes Welbl, Sven Gowal, Alexey Cherepanov, James Molloy,
  Daniel~J. Mankowitz, Esme~Sutherland Robson, Pushmeet Kohli, Nando
  de~Freitas, Koray Kavukcuoglu, and Oriol Vinyals.
\newblock Competition-level code generation with alphacode, 2022.
\newblock URL \url{https://arxiv.org/abs/2203.07814}.

\bibitem[Liang et~al.(2017)Liang, Berant, Le, Forbus, and Lao]{liang}
Chen Liang, Jonathan Berant, Quoc~V. Le, Ken Forbus, and Ni~Lao.
\newblock Neural symbolic machines: Learning semantic parsers on freebase with
  weak supervision.
\newblock In \emph{Proceedings of the 55th Annual Meeting of the Association
  for Computational Linguistics (Volume 1: Long Papers)}, pp.\  23--33,
  Vancouver, Canada, 2017.
\newblock URL
  \url{http://aclanthology.coli.uni-saarland.de/pdf/P/P17/P17-1003.pdf}.

\bibitem[Ni et~al.(2020)Ni, Yin, and Neubig]{ni2020merging}
Ansong Ni, Pengcheng Yin, and Graham Neubig.
\newblock Merging weak and active supervision for semantic parsing.
\newblock In \emph{Proceedings of the AAAI Conference on Artificial
  Intelligence}, volume~34, pp.\  8536--8543, 2020.

\bibitem[Ni et~al.(2022)Ni, Inala, Wang, Polozov, Meek, Radev, and Gao]{ni2022}
Ansong Ni, Jeevana~Priya Inala, Chenglong Wang, Oleksandr Polozov, Christopher
  Meek, Dragomir Radev, and Jianfeng Gao.
\newblock Learning from self-sampled correct and partially-correct programs,
  2022.
\newblock URL \url{https://arxiv.org/abs/2205.14318}.

\bibitem[Nijkamp et~al.(2022)Nijkamp, Pang, Hayashi, Tu, Wang, Zhou, Savarese,
  and Xiong]{nijkamp}
Erik Nijkamp, Bo~Pang, Hiroaki Hayashi, Lifu Tu, Huan Wang, Yingbo Zhou, Silvio
  Savarese, and Caiming Xiong.
\newblock A conversational paradigm for program synthesis, 2022.
\newblock URL \url{https://arxiv.org/abs/2203.13474}.

\bibitem[Papernot et~al.(2015)Papernot, McDaniel, Wu, Jha, and Swami]{papernot}
Nicolas Papernot, Patrick McDaniel, Xi~Wu, Somesh Jha, and Ananthram Swami.
\newblock Distillation as a defense to adversarial perturbations against deep
  neural networks, 2015.
\newblock URL \url{https://arxiv.org/abs/1511.04508}.

\bibitem[Pasupat \& Liang(2015)Pasupat and Liang]{pasupat2015compositional}
Panupong Pasupat and Percy Liang.
\newblock Compositional semantic parsing on semi-structured tables.
\newblock In \emph{Proceedings of the 53rd Annual Meeting of the Association
  for Computational Linguistics and the 7th International Joint Conference on
  Natural Language Processing (Volume 1: Long Papers)}, pp.\  1470--1480, 2015.

\bibitem[Polu et~al.(2022)Polu, Han, Zheng, Baksys, Babuschkin, and
  Sutskever]{polu}
Stanislas Polu, Jesse~Michael Han, Kunhao Zheng, Mantas Baksys, Igor
  Babuschkin, and Ilya Sutskever.
\newblock Formal mathematics statement curriculum learning, 2022.
\newblock URL \url{https://arxiv.org/abs/2202.01344}.

\bibitem[Sanh et~al.(2019)Sanh, Debut, Chaumond, and Wolf]{sanh}
Victor Sanh, Lysandre Debut, Julien Chaumond, and Thomas Wolf.
\newblock Distilbert, a distilled version of bert: smaller, faster, cheaper and
  lighter, 2019.
\newblock URL \url{https://arxiv.org/abs/1910.01108}.

\bibitem[Silver et~al.(2017)Silver, Hubert, Schrittwieser, Antonoglou, Lai,
  Guez, Lanctot, Sifre, Kumaran, Graepel, Lillicrap, Simonyan, and
  Hassabis]{silver}
David Silver, Thomas Hubert, Julian Schrittwieser, Ioannis Antonoglou, Matthew
  Lai, Arthur Guez, Marc Lanctot, Laurent Sifre, Dharshan Kumaran, Thore
  Graepel, Timothy Lillicrap, Karen Simonyan, and Demis Hassabis.
\newblock Mastering chess and shogi by self-play with a general reinforcement
  learning algorithm, 2017.
\newblock URL \url{https://arxiv.org/abs/1712.01815}.

\bibitem[Thoppilan et~al.(2022)Thoppilan, De~Freitas, Hall, Shazeer,
  Kulshreshtha, Cheng, Jin, Bos, Baker, Du, Li, Lee, Zheng, Ghafouri, Menegali,
  Huang, Krikun, Lepikhin, Qin, Chen, Xu, Chen, Roberts, Bosma, Zhao, Zhou,
  Chang, Krivokon, Rusch, Pickett, Srinivasan, Man, Meier-Hellstern, Morris,
  Doshi, Santos, Duke, Soraker, Zevenbergen, Prabhakaran, Diaz, Hutchinson,
  Olson, Molina, Hoffman-John, Lee, Aroyo, Rajakumar, Butryna, Lamm, Kuzmina,
  Fenton, Cohen, Bernstein, Kurzweil, Aguera-Arcas, Cui, Croak, Chi, and
  Le]{lamda}
Romal Thoppilan, Daniel De~Freitas, Jamie Hall, Noam Shazeer, Apoorv
  Kulshreshtha, Heng-Tze Cheng, Alicia Jin, Taylor Bos, Leslie Baker, Yu~Du,
  YaGuang Li, Hongrae Lee, Huaixiu~Steven Zheng, Amin Ghafouri, Marcelo
  Menegali, Yanping Huang, Maxim Krikun, Dmitry Lepikhin, James Qin, Dehao
  Chen, Yuanzhong Xu, Zhifeng Chen, Adam Roberts, Maarten Bosma, Vincent Zhao,
  Yanqi Zhou, Chung-Ching Chang, Igor Krivokon, Will Rusch, Marc Pickett,
  Pranesh Srinivasan, Laichee Man, Kathleen Meier-Hellstern, Meredith~Ringel
  Morris, Tulsee Doshi, Renelito~Delos Santos, Toju Duke, Johnny Soraker, Ben
  Zevenbergen, Vinodkumar Prabhakaran, Mark Diaz, Ben Hutchinson, Kristen
  Olson, Alejandra Molina, Erin Hoffman-John, Josh Lee, Lora Aroyo, Ravi
  Rajakumar, Alena Butryna, Matthew Lamm, Viktoriya Kuzmina, Joe Fenton, Aaron
  Cohen, Rachel Bernstein, Ray Kurzweil, Blaise Aguera-Arcas, Claire Cui,
  Marian Croak, Ed~Chi, and Quoc Le.
\newblock Lamda: Language models for dialog applications, 2022.
\newblock URL \url{https://arxiv.org/abs/2201.08239}.

\bibitem[Wang \& Komatsuzaki(2021)Wang and Komatsuzaki]{gpt-j}
Ben Wang and Aran Komatsuzaki.
\newblock {GPT-J-6B: A 6 Billion Parameter Autoregressive Language Model}.
\newblock \url{https://github.com/kingoflolz/mesh-transformer-jax}, May 2021.

\bibitem[Wei et~al.(2022)Wei, Tay, Bommasani, Raffel, Zoph, Borgeaud, Yogatama,
  Bosma, Zhou, Metzler, Chi, Hashimoto, Vinyals, Liang, Dean, and
  Fedus]{wei2022emergent}
Jason Wei, Yi~Tay, Rishi Bommasani, Colin Raffel, Barret Zoph, Sebastian
  Borgeaud, Dani Yogatama, Maarten Bosma, Denny Zhou, Donald Metzler, Ed~H.
  Chi, Tatsunori Hashimoto, Oriol Vinyals, Percy Liang, Jeff Dean, and William
  Fedus.
\newblock Emergent abilities of large language models, 2022.

\bibitem[Xu et~al.(2022)Xu, Alon, Neubig, and Hellendoorn]{xu2022systematic}
Frank~F Xu, Uri Alon, Graham Neubig, and Vincent~J Hellendoorn.
\newblock A systematic evaluation of large language models of code.
\newblock \emph{arXiv preprint arXiv:2202.13169}, 2022.

\bibitem[Zelle \& Mooney(1996)Zelle and Mooney]{zelle}
John~M. Zelle and Raymond~J. Mooney.
\newblock Learning to parse database queries using inductive logic programming.
\newblock In \emph{Proceedings of the Thirteenth National Conference on
  Artificial Intelligence - Volume 2}, AAAI'96, pp.\  1050–1055. AAAI Press,
  1996.
\newblock ISBN 026251091X.

\bibitem[Zettlemoyer \& Collins(2012)Zettlemoyer and Collins]{zettlemoyer}
Luke~S. Zettlemoyer and Michael Collins.
\newblock Learning to map sentences to logical form: Structured classification
  with probabilistic categorial grammars, 2012.
\newblock URL \url{https://arxiv.org/abs/1207.1420}.

\bibitem[Zhong et~al.(2017)Zhong, Xiong, and Socher]{zhong}
Victor Zhong, Caiming Xiong, and Richard Socher.
\newblock Seq2sql: Generating structured queries from natural language using
  reinforcement learning, 2017.
\newblock URL \url{https://arxiv.org/abs/1709.00103}.

\end{thebibliography}
